\documentclass[10pt,conference]{IEEEtran}

\usepackage[utf8]{inputenc}
\usepackage[T1]{fontenc}
\usepackage{amsmath,amssymb,amsfonts}
\usepackage{graphicx}
\usepackage{booktabs}
\usepackage{array}
\usepackage{multirow}
\usepackage{xcolor}
\usepackage{hyperref}
\usepackage{algorithm}
\usepackage{algorithmic}
\usepackage{tikz}
\usetikzlibrary{shapes,arrows,positioning,fit,backgrounds,calc,decorations.pathreplacing,matrix,patterns,shadows}
\usepackage{pgfplots}
\pgfplotsset{compat=1.17}
\usepackage{caption}
\usepackage{subcaption}
\usepackage{cite}
\usepackage{balance}
\usepackage{colortbl}
\usepackage{tabularx}

\definecolor{timesfm}{RGB}{66,133,244}
\definecolor{attention}{RGB}{219,68,55}
\definecolor{bilinear}{RGB}{15,157,88}
\definecolor{output}{RGB}{244,180,0}
\definecolor{premarket}{RGB}{171,71,188}
\definecolor{fusion}{RGB}{255,112,67}
\definecolor{rf}{RGB}{0,150,136}
\definecolor{positive}{RGB}{76,175,80}
\definecolor{negative}{RGB}{244,67,54}
\definecolor{neutral}{RGB}{158,158,158}
\definecolor{highlight}{RGB}{255,235,59}
\definecolor{darkblue}{RGB}{25,118,210}
\definecolor{lightgray}{RGB}{245,245,245}
\definecolor{revcolor}{RGB}{0,0,200}

\newcommand{\rev}[1]{#1}

\title{Hybrid Neural-Classical Correction for Frozen Time Series Foundation Models: A Comprehensive Ablation Study on High-Frequency Stock Prediction}

\author{
\IEEEauthorblockN{1\textsuperscript{st} Kasun Dewage}
\IEEEauthorblockA{\textit{Dept. of Mathematics} \\
\textit{University of Central Florida}\\
Orlando, FL, USA \\
KasunTharuka.Dewage@ucf.edu}
\and
\IEEEauthorblockN{2\textsuperscript{nd} Suranadi De Silva}
\IEEEauthorblockA{\textit{Dept. of Computer Science} \\
\textit{University of Central Florida}\\
Orlando, FL, USA \\
su966204@ucf.edu}
\and
\IEEEauthorblockN{3\textsuperscript{rd} Shankhadeep Mondal}
\IEEEauthorblockA{\textit{Dept. of Mathematics} \\
\textit{University of Central Florida}\\
Orlando, FL, USA \\
shankhadeep.mondal@ucf.edu}
}

\begin{document}

\maketitle

\begin{tikzpicture}[remember picture,overlay]
\node[anchor=south,align=center,text width=7.1in,font=\fontsize{5.5}{6.2}\selectfont]
  at ([yshift=0.16in]current page.south)
  {\copyright~2026 IEEE. Personal use of this material is permitted. Permission from IEEE must be obtained for all other uses, in any current or future media, including reprinting/republishing this material for advertising or promotional purposes, creating new collective works, for resale or redistribution to servers or lists, or reuse of any copyrighted component of this work in other works.};
\end{tikzpicture}

\begin{abstract}
Foundation models for time series forecasting demonstrate impressive zero-shot generalization but often underperform on specialized domains such as high-frequency finance. We present a comprehensive study of \textit{hybrid neural-classical correction} for adapting frozen TimesFM (200M parameters) to stock return prediction during the volatile opening trading hour. We compare two neural correction architectures---\textbf{AttnCorrect} (multi-head self-attention, ${\sim}$471K parameters) and \textbf{GatedLinear} (low-rank bilinear projection with gating, ${\sim}$49K parameters)---each augmented with Random Forest residual learning. Through systematic ablation across \textbf{10 major technology stocks} (NVDA, MSFT, AAPL, GOOG, GOOGL, AMZN, META, AVGO, TSLA, NFLX) spanning 2 million data points, we reveal critical insights: (1)~The hybrid neural-classical approach achieves \textbf{\rev{0.597 pooled correlation and 6.4$\times$ mean per-day correlation improvement}} over frozen TimesFM; (2)~Classical residual learning (Random Forest) \rev{provides the largest single-component contribution }, \textbf{\rev{ matching or exceeding}} the neural correction component; (3)~Simpler neural architectures surprisingly outperform complex ones when classical residual learning is removed; (4)~Self-attention provides the largest \rev{ neural-only} contribution. GatedLinear+RF achieves best overall performance  with \textbf{9$\times$ fewer neural parameters} than AttnCorrect+RF. \rev{We report three complementary correlation metrics---mean per-day, cross-day cumulative, and pooled---to provide a complete picture of predictive quality.} Our results provide practical guidance: effective foundation model adaptation requires careful integration of neural and classical components, with classical methods playing a crucial complementary role.
\end{abstract}


\section{Introduction}

Foundation models have revolutionized machine learning across domains. Recent extensions to time series---including TimesFM \cite{das2024decoder}, Chronos \cite{ansari2024chronos}, and Lag-Llama \cite{rasul2024lagllama}---demonstrate that models pretrained on billions of time points can achieve strong zero-shot generalization across diverse forecasting tasks.

However, zero-shot performance does not guarantee domain-specific accuracy. When applying TimesFM to high-frequency stock prediction, we observe \textbf{near-zero \rev{mean per-day} correlation} (\rev{0.059}) with actual returns---essentially uninformative predictions. This gap between general capability and domain-specific performance motivates investigation into effective adaptation strategies.

We focus on a challenging prediction task: forecasting stock returns during the \textbf{opening trading hour} (9:30--10:30 AM), when markets exhibit high volatility as they digest overnight information \cite{hasbrouck2013low}. Premarket trading data (4:30--9:29 AM) provides potentially predictive signals, but extracting useful information requires sophisticated processing.

Rather than fine-tuning the foundation model---computationally expensive and prone to overfitting with limited domain data---we investigate \textbf{correction-based adaptation}: keeping TimesFM completely frozen while training lightweight modules to correct its outputs. Critically, we employ a \textbf{hybrid neural-classical design} combining neural correction with Random Forest \cite{breiman2001random} residual learning.

We compare two neural correction architectures representing different design philosophies:

\begin{itemize}
    \item \textbf{AttnCorrect}: Multi-head self-attention \cite{vaswani2017attention} over premarket sequences, enabling flexible temporal pattern learning (${\sim}$471K trainable parameters)
    \item \textbf{GatedLinear}: Low-rank bilinear projection \cite{tenenbaum2000separating} with learned gating for extreme parameter efficiency (${\sim}$49K trainable parameters)
\end{itemize}

Through comprehensive ablation across \textbf{10 major technology stocks} with over 2 million data points, we address the critical question:

\begin{center}
\textit{What components matter most when adapting frozen foundation models, and how should neural and classical methods be integrated?}
\end{center}

\textbf{Our key contributions and findings:}

\begin{enumerate}
    \item A \textbf{hybrid neural-classical framework} achieving \rev{6.4$\times$ mean per-day correlation improvement and 0.597 pooled correlation} over frozen TimesFM
    \item Comprehensive \textbf{ablation across 12 model variants} revealing component contributions
    \item Evidence that \textbf{classical residual learning \rev{matches or exceeding neural correction}} 
    \item The surprising finding that \textbf{simpler neural architectures outperform complex ones} when classical components are removed
    \item \textbf{Per-stock analysis} across 10 major technology stocks with detailed performance breakdowns
    \item Practical guidance: GatedLinear+RF achieves best performance with 9$\times$ fewer neural parameters
\end{enumerate}

\noindent\textbf{Open Science Statement:} Our code and results  are publicly available at \url{https://github.com/Kasun-Dewage/Hybrid_Neural2026.git} to facilitate reproducibility and further research in this area upon publication of this manuscript. All experiments use a fixed random seed of 42 for reproducibility.

\section{Related Work}

\subsection{Foundation Models for Time Series}

TimesFM \cite{das2024decoder} is a decoder-only transformer with 200M parameters, pretrained on over 100 billion time points from Google Trends, Wikipedia pageviews, and synthetic data. The model uses input patching with patch length 32 and achieves strong zero-shot performance on standard benchmarks including ETT, Weather, and Electricity datasets.

Chronos \cite{ansari2024chronos} takes a different approach, tokenizing time series values into discrete bins and training T5-style encoder-decoder models \cite{raffel2020exploring}. This enables the use of language modeling techniques for time series. Lag-Llama \cite{rasul2024lagllama} adapts the LLaMA architecture with lag-based tokenization for probabilistic forecasting, demonstrating that decoder-only architectures can effectively model temporal dependencies.

While these models generalize impressively to standard benchmarks, domain adaptation to specialized applications like high-frequency finance remains challenging due to the unique statistical properties of financial time series.

\subsection{Efficient Model Adaptation}

Parameter-efficient fine-tuning has been extensively studied for large language models. LoRA \cite{hu2022lora} introduces low-rank decomposition of weight updates, training only $\mathbf{W} + \mathbf{BA}$ where $\mathbf{B} \in \mathbb{R}^{d \times r}$ and $\mathbf{A} \in \mathbb{R}^{r \times k}$ with $r \ll \min(d,k)$. Adapter layers \cite{houlsby2019parameter} insert small bottleneck modules between transformer layers. Prefix tuning \cite{li2021prefix} prepends learnable tokens to the input sequence.

These methods modify model internals, requiring access to architecture details and intermediate representations. Our approach operates \textit{externally} through output correction, treating the foundation model as a black box. This enables adaptation without architectural knowledge and avoids potential instabilities from modifying pretrained weights.

\subsection{Hybrid Neural-Classical Methods}

The integration of neural networks with classical machine learning methods has shown success across domains. In forecasting, hybrid approaches combining neural networks with statistical methods often outperform pure neural approaches \cite{makridakis2018statistical}. Recent work demonstrates that gradient boosting and Random Forests remain highly competitive with deep learning for tabular data \cite{grinsztajn2022tree}, particularly when feature engineering captures domain knowledge.

Random Forests \cite{breiman2001random} offer several advantages for residual learning: robustness to outliers, natural handling of feature interactions, and strong performance with limited training data. Our work provides empirical evidence for the value of neural-classical integration in foundation model adaptation.

\subsection{Attention Mechanisms and Bilinear Models}

The Transformer architecture \cite{vaswani2017attention} introduced scaled dot-product attention, enabling flexible modeling of long-range dependencies. For time series, attention has been adapted in various ways: the Temporal Fusion Transformer \cite{lim2021temporal} combines LSTM encoders with multi-head attention for interpretable forecasting, while Informer \cite{zhou2021informer} introduces ProbSparse attention for computational efficiency.

Bilinear models capture multiplicative interactions between features \cite{tenenbaum2000separating}. Low-rank bilinear pooling \cite{kim2017hadamard} reduces computation via matrix factorization. We apply bilinear projection to compress high-dimensional premarket sequences with minimal parameters.

\subsection{Financial Time Series Prediction}

Despite efficient market hypothesis arguments \cite{fama1970efficient}, empirical evidence supports short-term predictability during information asymmetry periods. Deep learning approaches using LSTM \cite{fischer2018deep} and attention mechanisms have shown promise for financial forecasting. The opening trading hour exhibits particularly high volatility and potential predictability as markets process overnight information \cite{hasbrouck2013low}.

\section{Methodology}

\subsection{Problem Setting}

Let $\mathbf{X}^{(pm)} \in \mathbb{R}^{T \times F}$ denote the premarket sequence with $T = 300$ one-minute bars and $F = 7$ features:
\begin{itemize}
    \item Normalized OHLC prices (4 features)
    \item Log-transformed volume (1 feature)
    \item Bar-to-bar momentum (1 feature)
    \item Intrabar volatility (1 feature)
\end{itemize}

Given frozen foundation model prediction $\hat{\mathbf{y}}^{(tfm)} \in \mathbb{R}^{H}$ for horizon $H = 60$ minutes and multiscale summary features $\mathbf{m} \in \mathbb{R}^{21}$, we learn a hybrid correction:

\begin{equation}
    \hat{\mathbf{y}} = \hat{\mathbf{y}}^{(tfm)} + \underbrace{\Delta\mathbf{y}_{neural}(\mathbf{X}^{(pm)}, \mathbf{m}, \hat{\mathbf{y}}^{(tfm)})}_{\text{Neural correction}} + \underbrace{\Delta\mathbf{y}_{RF}(f_{summary})}_{\text{Classical residual}}
\end{equation}

The foundation model remains completely frozen throughout; only the correction modules are trained.

\subsection{Hybrid Correction Framework}

\begin{figure}[t]
\centering
\resizebox{0.98\columnwidth}{!}{
\begin{tikzpicture}[
    node distance=0.4cm,
    box/.style={rectangle, draw, rounded corners=3pt, minimum width=2cm, minimum height=0.52cm, align=center, font=\small, drop shadow={shadow xshift=0.7pt, shadow yshift=-0.7pt, opacity=0.22}},
    bigbox/.style={rectangle, draw, rounded corners=4pt, minimum width=2.4cm, minimum height=0.62cm, align=center, font=\small\bfseries, drop shadow={shadow xshift=0.7pt, shadow yshift=-0.7pt, opacity=0.22}},
    smallbox/.style={rectangle, draw, rounded corners=2pt, minimum width=1.5cm, minimum height=0.42cm, align=center, font=\scriptsize},
    arrow/.style={->, >=stealth, thick, draw=gray!65},
    highlight_arrow/.style={->, >=stealth, very thick, draw=rf!75},
]

\node[box, fill=blue!15] (context) {Context Sequence\\(Prev day + PM)};
\node[box, fill=premarket!22, right=0.9cm of context] (premarket) {Premarket Tensor\\$\mathbf{X}^{(pm)} \in \mathbb{R}^{300 \times 7}$};
\node[box, fill=blue!15, right=0.9cm of premarket] (multiscale) {Multiscale Features\\$\mathbf{m} \in \mathbb{R}^{21}$};

\node[bigbox, fill=timesfm!28, below=0.65cm of context] (timesfm) {\textbf{TimesFM (200M)}\\Frozen Foundation};

\node[bigbox, fill=attention!22, below=0.65cm of premarket, minimum width=3.6cm] (neural) {\textbf{Neural Corrector}\\(AttnCorrect or GatedLinear)};

\node[smallbox, fill=timesfm!16, below=0.42cm of timesfm] (tfm_out) {$\hat{\mathbf{y}}^{(tfm)} \in \mathbb{R}^{60}$};
\node[smallbox, fill=attention!16, below=0.42cm of neural] (neural_out) {$\Delta\mathbf{y}_{neural} \in \mathbb{R}^{60}$};

\node[box, fill=purple!20, below=0.6cm of neural_out, xshift=-0.85cm] (add1) {$\hat{\mathbf{y}}^{(tfm)} + \Delta\mathbf{y}_{neural}$};

\node[bigbox, fill=rf!42, line width=1.8pt, draw=rf!75, below=0.42cm of add1] (rf) {\textbf{Random Forest}\\250 trees, depth 12};

\node[bigbox, fill=output!32, below=0.42cm of rf] (final) {\textbf{Final Prediction}\\$\hat{\mathbf{y}}_{final} \in \mathbb{R}^{60}$};

\draw[arrow] (context) -- (timesfm);
\draw[arrow] (premarket) -- (neural);
\draw[arrow] (multiscale) |- ([yshift=0.22cm]neural.east) -- (neural.east);
\draw[arrow] (timesfm) -- (tfm_out);
\draw[arrow] (neural) -- (neural_out);
\draw[arrow] (tfm_out) |- (add1);
\draw[arrow] (neural_out) -- (add1);
\draw[arrow] (tfm_out) -- ++(0,-0.22) -| ([xshift=-0.35cm]neural.south);
\draw[highlight_arrow] (add1) -- (rf);
\draw[highlight_arrow] (rf) -- (final);

\node[right=0.06cm of rf, font=\scriptsize\bfseries, text=rf, text width=1.8cm, align=left] {\textbf{\rev{+0.16 Corr}}\\Classical\\Residual};
\node[left=0.05cm of timesfm, font=\tiny, text=gray, rotate=90] {FROZEN};
\node[left=0.05cm of neural, font=\tiny, text=gray, rotate=90] {TRAINED};

\end{tikzpicture}
}
\caption{\textbf{Hybrid Neural-Classical Correction Framework.} Both architectures share this pipeline: frozen TimesFM provides base predictions, a trainable neural corrector generates $\Delta\mathbf{y}_{neural}$, and Random Forest learns residual patterns. Our ablation reveals the classical component (RF, highlighted) \rev{provides nearly equally or exceeding to the neural component}.}
\label{fig:framework}
\end{figure}
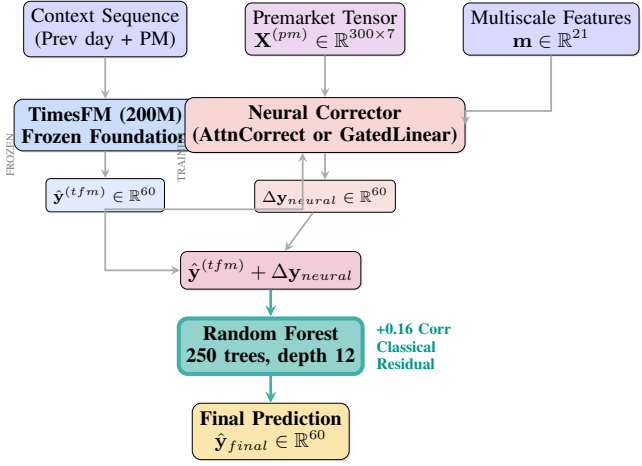

Figure \ref{fig:framework} illustrates the complete hybrid neural-classical correction framework shared by both architectures.

Both architectures share these components:
\begin{itemize}
    \item \textbf{TimesFM backbone}: Frozen 200M-parameter foundation model
    \item \textbf{Multiscale features $\mathbf{m}$}: Returns, volatility, and volume computed over 5/15/30/60-minute windows plus overnight gap (21 dimensions total)
    \item \textbf{Random Forest residual}: 250 trees, max depth 12, min samples leaf 3, trained on neural correction residuals
    \item \textbf{Directional loss}: $\mathcal{L} = \mathcal{L}_{MSE} + 0.3\mathcal{L}_{dir} + 0.2\mathcal{L}_{cum}$ to encourage correct direction prediction
\end{itemize}

\subsection{AttnCorrect: Attention-Based Correction}

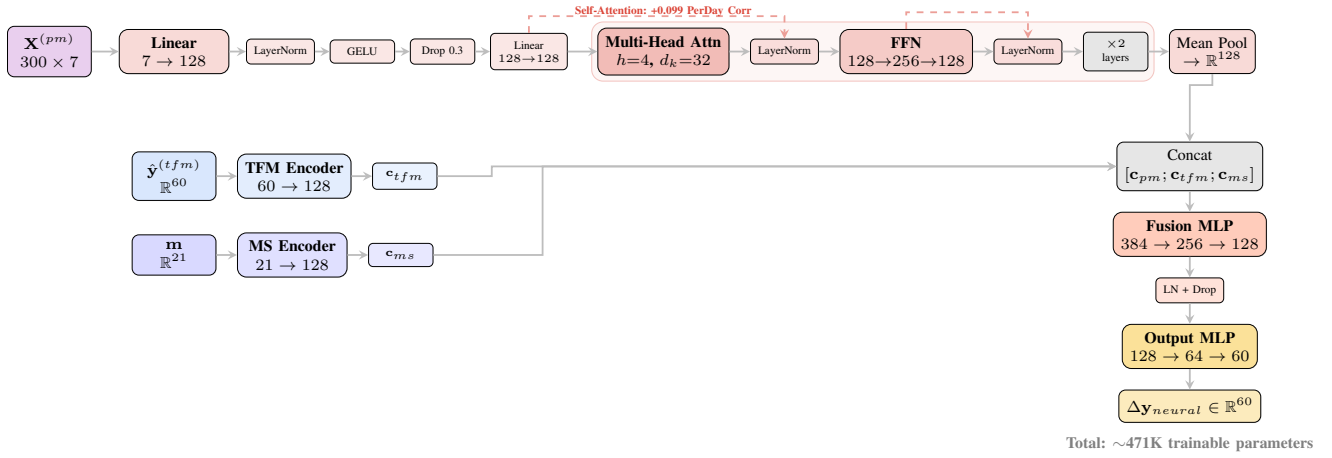
\begin{figure*}[t]
\centering
\resizebox{0.97\textwidth}{!}{
\begin{tikzpicture}[
    node distance=0.32cm,
    box/.style={rectangle, draw, rounded corners=3pt, minimum width=1.3cm, minimum height=0.46cm, align=center, font=\scriptsize, drop shadow={shadow xshift=0.35pt, shadow yshift=-0.35pt, opacity=0.16}},
    smallbox/.style={rectangle, draw, rounded corners=2pt, minimum width=1cm, minimum height=0.36cm, align=center, font=\tiny},
    bigbox/.style={rectangle, draw, rounded corners=4pt, minimum width=1.7cm, minimum height=0.5cm, align=center, font=\scriptsize\bfseries, drop shadow={shadow xshift=0.35pt, shadow yshift=-0.35pt, opacity=0.16}},
    arrow/.style={->, >=stealth, thick, draw=gray!52},
    residual/.style={->, >=stealth, thick, dashed, draw=attention!52},
]

\node[box, fill=premarket!26] (input) {$\mathbf{X}^{(pm)}$\\$300 \times 7$};

\node[bigbox, fill=attention!20, right=0.42cm of input] (embed) {Linear\\$7 \to 128$};
\node[smallbox, fill=attention!13, right=0.26cm of embed] (ln1) {LayerNorm};
\node[smallbox, fill=attention!13, right=0.23cm of ln1] (gelu1) {GELU};
\node[smallbox, fill=attention!13, right=0.23cm of gelu1] (drop1) {Drop 0.3};
\node[smallbox, fill=attention!13, right=0.23cm of drop1] (lin2) {Linear\\$128{\to}128$};

\node[bigbox, fill=attention!36, right=0.46cm of lin2, minimum width=2cm] (mha) {Multi-Head Attn\\$h{=}4$, $d_k{=}32$};
\node[smallbox, fill=attention!18, right=0.32cm of mha] (ln2) {LayerNorm};
\node[bigbox, fill=attention!28, right=0.32cm of ln2, minimum width=1.85cm] (ffn) {FFN\\$128{\to}256{\to}128$};
\node[smallbox, fill=attention!18, right=0.32cm of ffn] (ln3) {LayerNorm};
\node[smallbox, fill=gray!20, right=0.32cm of ln3] (repeat) {$\times 2$\\layers};
\node[box, fill=attention!20, right=0.32cm of repeat] (pool) {Mean Pool\\$\to \mathbb{R}^{128}$};

\node[box, fill=timesfm!20, below=1.2cm of embed] (tfm_in) {$\hat{\mathbf{y}}^{(tfm)}$\\$\mathbb{R}^{60}$};
\node[bigbox, fill=timesfm!15, right=0.32cm of tfm_in] (tfm_enc) {TFM Encoder\\$60 \to 128$};
\node[smallbox, fill=timesfm!11, right=0.32cm of tfm_enc] (tfm_out) {$\mathbf{c}_{tfm}$};

\node[box, fill=blue!15, below=0.52cm of tfm_in] (ms_in) {$\mathbf{m}$\\$\mathbb{R}^{21}$};
\node[bigbox, fill=blue!12, right=0.32cm of ms_in] (ms_enc) {MS Encoder\\$21 \to 128$};
\node[smallbox, fill=blue!9, right=0.32cm of ms_enc] (ms_out) {$\mathbf{c}_{ms}$};

\node[box, fill=gray!20, below=1.05cm of pool, xshift=-0.35cm] (concat) {Concat\\$[\mathbf{c}_{pm}; \mathbf{c}_{tfm}; \mathbf{c}_{ms}]$};
\node[bigbox, fill=fusion!36, below=0.32cm of concat] (fusion) {Fusion MLP\\$384 \to 256 \to 128$};
\node[smallbox, fill=fusion!20, below=0.32cm of fusion] (ln_fuse) {LN + Drop};

\node[bigbox, fill=output!40, below=0.32cm of ln_fuse] (output) {Output MLP\\$128 \to 64 \to 60$};
\node[box, fill=output!26, below=0.32cm of output] (delta) {$\Delta\mathbf{y}_{neural} \in \mathbb{R}^{60}$};

\draw[arrow] (input) -- (embed);
\draw[arrow] (embed) -- (ln1);
\draw[arrow] (ln1) -- (gelu1);
\draw[arrow] (gelu1) -- (drop1);
\draw[arrow] (drop1) -- (lin2);
\draw[arrow] (lin2) -- (mha);
\draw[arrow] (mha) -- (ln2);
\draw[arrow] (ln2) -- (ffn);
\draw[arrow] (ffn) -- (ln3);
\draw[arrow] (ln3) -- (repeat);
\draw[arrow] (repeat) -- (pool);

\draw[arrow] (tfm_in) -- (tfm_enc);
\draw[arrow] (tfm_enc) -- (tfm_out);
\draw[arrow] (ms_in) -- (ms_enc);
\draw[arrow] (ms_enc) -- (ms_out);

\draw[arrow] (pool) -- ++(0,-0.52) -| (concat);
\draw[arrow] (tfm_out) -- ++(1.35,0) |- (concat);
\draw[arrow] (ms_out) -- ++(2.2,0) |- (concat);

\draw[arrow] (concat) -- (fusion);
\draw[arrow] (fusion) -- (ln_fuse);
\draw[arrow] (ln_fuse) -- (output);
\draw[arrow] (output) -- (delta);

\draw[residual] (lin2.north) -- ++(0,0.26) -| (ln2.north);
\draw[residual] (ffn.north) -- ++(0,0.26) -| (ln3.north);

\node[below=0.1cm of delta, font=\scriptsize, text=gray] {\textbf{Total: ${\sim}$471K trainable parameters}};
\node[above=0.05cm of mha, font=\tiny\bfseries, text=attention] {Self-Attention: \rev{+0.099 PerDay Corr}};

\begin{scope}[on background layer]
\node[draw=attention!32, fill=attention!5, rounded corners=5pt, fit=(mha)(ln2)(ffn)(ln3)(repeat), inner sep=2.5pt] {};
\end{scope}

\end{tikzpicture}
}
\caption{\textbf{AttnCorrect Architecture.} Premarket sequences ($300 \times 7$) pass through embedding layers, then $L{=}2$ transformer blocks with 4-head self-attention (highlighted box). Pooled features fuse with encoded TimesFM predictions and multiscale features. Output weights initialized to zero \cite{zhang2019fixup} for stable training.}
\label{fig:attn_arch}
\end{figure*}

Figure \ref{fig:attn_arch} presents the complete AttnCorrect architecture with ${\sim}$471K trainable parameters.

\subsubsection{Premarket Encoding}

The sequence is projected to hidden dimension $d = 128$:
\begin{equation}
\mathbf{H}^{(0)} = \text{Linear}_{128}(\text{Drop}_{0.3}(\text{GELU}(\text{LN}(\mathbf{X}^{(pm)} \mathbf{W}_e + \mathbf{b}_e))))
\end{equation}
where $\mathbf{W}_e \in \mathbb{R}^{7 \times 128}$. GELU \cite{hendrycks2016gaussian} provides smooth non-linearity and LayerNorm \cite{ba2016layer} stabilizes training.

\subsubsection{Self-Attention Layers}

We apply $L = 2$ transformer layers with $h = 4$ heads and head dimension $d_k = 32$:
\begin{equation}
    \mathbf{H}' = \text{LN}(\mathbf{H}^{(\ell-1)} + \text{MHA}(\mathbf{H}^{(\ell-1)}))
\end{equation}
\begin{equation}
    \mathbf{H}^{(\ell)} = \text{LN}(\mathbf{H}' + \text{FFN}(\mathbf{H}'))
\end{equation}

The feed-forward network uses expansion factor 2:
\begin{equation}
    \text{FFN}(\mathbf{x}) = \text{Drop}_{0.3}(\text{GELU}(\mathbf{x}\mathbf{W}_1))\mathbf{W}_2
\end{equation}
with $\mathbf{W}_1 \in \mathbb{R}^{128 \times 256}$ and $\mathbf{W}_2 \in \mathbb{R}^{256 \times 128}$.

\subsubsection{Cross-Modal Fusion}

The pooled context $\mathbf{c}_{pm} = \frac{1}{T}\sum_t \mathbf{H}^{(L)}_t$ is concatenated with separately encoded TimesFM predictions and multiscale features:
\begin{equation}
    \mathbf{h} = \text{MLP}_{fusion}([\mathbf{c}_{pm}; \mathbf{c}_{tfm}; \mathbf{c}_{ms}])
\end{equation}
where the fusion MLP maps $\mathbb{R}^{384} \to \mathbb{R}^{128}$.

\subsection{GatedLinear: Bilinear-Gated Correction}

\begin{figure*}[t]
\centering
\resizebox{0.97\textwidth}{!}{
\begin{tikzpicture}[
    node distance=0.32cm,
    box/.style={rectangle, draw, rounded corners=3pt, minimum width=1.3cm, minimum height=0.46cm, align=center, font=\scriptsize, drop shadow={shadow xshift=0.35pt, shadow yshift=-0.35pt, opacity=0.16}},
    smallbox/.style={rectangle, draw, rounded corners=2pt, minimum width=1cm, minimum height=0.36cm, align=center, font=\tiny},
    bigbox/.style={rectangle, draw, rounded corners=4pt, minimum width=1.7cm, minimum height=0.5cm, align=center, font=\scriptsize\bfseries, drop shadow={shadow xshift=0.35pt, shadow yshift=-0.35pt, opacity=0.16}},
    arrow/.style={->, >=stealth, thick, draw=gray!52},
]

\node[box, fill=premarket!26] (input) {$\mathbf{X}^{(pm)}$\\$300 \times 7$};

\node[bigbox, fill=bilinear!30, right=0.52cm of input, minimum width=2cm] (V) {Temporal Proj.\\$\mathbf{V} \in \mathbb{R}^{300 \times 8}$};
\node[smallbox, fill=bilinear!18, right=0.32cm of V] (VX) {$\mathbf{V}^\top\mathbf{X}$\\$8 \times 7$};
\node[bigbox, fill=bilinear!30, right=0.32cm of VX, minimum width=1.85cm] (U) {Feature Proj.\\$\mathbf{U} \in \mathbb{R}^{7 \times 4}$};
\node[smallbox, fill=bilinear!18, right=0.32cm of U] (Z) {$\mathbf{Z}$\\$8 \times 4$};
\node[box, fill=bilinear!22, right=0.32cm of Z] (flatten) {Flatten\\$\to \mathbb{R}^{32}$};

\node[box, fill=timesfm!20, below=1.05cm of V] (tfm_in) {$\hat{\mathbf{y}}^{(tfm)}$\\$\mathbb{R}^{60}$};
\node[box, fill=blue!15, below=0.52cm of tfm_in] (ms_in) {$\mathbf{m}$\\$\mathbb{R}^{21}$};

\node[box, fill=gray!20, right=0.65cm of flatten, yshift=-1.15cm] (concat) {Concat\\$[\mathbf{z}; \mathbf{m}; \hat{\mathbf{y}}^{(tfm)}]$\\$\mathbb{R}^{113}$};

\node[bigbox, fill=bilinear!26, below=0.42cm of concat] (enc1) {Linear $113 \to 128$\\LayerNorm + GELU};
\node[smallbox, fill=bilinear!16, below=0.3cm of enc1] (drop) {Dropout 0.25};
\node[bigbox, fill=bilinear!26, below=0.3cm of drop] (enc2) {Linear $128 \to 128$\\GELU};

\node[box, fill=output!46, below left=0.62cm and 0.22cm of enc2] (gate_lin) {Gate Linear\\$128 \to 60$};
\node[smallbox, fill=output!56, below=0.26cm of gate_lin] (sigmoid) {Sigmoid};
\node[box, fill=output!36, below=0.26cm of sigmoid] (g) {$\mathbf{g} \in [0,1]^{60}$};

\node[box, fill=bilinear!36, below right=0.62cm and 0.22cm of enc2] (res_lin) {Residual Linear\\$128 \to 60$};
\node[box, fill=bilinear!26, below=0.58cm of res_lin] (r) {$\mathbf{r} \in \mathbb{R}^{60}$};

\node[bigbox, fill=output!30, below=1.98cm of enc2] (multiply) {$\Delta\mathbf{y} = \mathbf{g} \odot \mathbf{r}$};
\node[box, fill=output!20, below=0.32cm of multiply] (delta) {$\Delta\mathbf{y}_{neural} \in \mathbb{R}^{60}$};

\draw[arrow] (input) -- (V);
\draw[arrow] (V) -- (VX);
\draw[arrow] (VX) -- (U);
\draw[arrow] (U) -- (Z);
\draw[arrow] (Z) -- (flatten);

\draw[arrow] (flatten) -- ++(0,-0.72) -| (concat);
\draw[arrow] (tfm_in) -| (concat);
\draw[arrow] (ms_in) -| (concat);

\draw[arrow] (concat) -- (enc1);
\draw[arrow] (enc1) -- (drop);
\draw[arrow] (drop) -- (enc2);

\draw[arrow] (enc2) -| (gate_lin);
\draw[arrow] (enc2) -| (res_lin);
\draw[arrow] (gate_lin) -- (sigmoid);
\draw[arrow] (sigmoid) -- (g);
\draw[arrow] (res_lin) -- (r);
\draw[arrow] (g) |- (multiply);
\draw[arrow] (r) |- (multiply);
\draw[arrow] (multiply) -- (delta);

\node[below=0.1cm of delta, font=\scriptsize, text=gray] {\textbf{Total: ${\sim}$49K params (9$\times$ fewer than AttnCorrect)}};
\node[above=0.05cm of V, font=\tiny, text=bilinear] {Bilinear: only 2,428 params};

\begin{scope}[on background layer]
\node[draw=bilinear!32, fill=bilinear!5, rounded corners=5pt, fit=(V)(VX)(U)(Z), inner sep=2.5pt] {};
\node[draw=output!32, fill=output!5, rounded corners=5pt, fit=(gate_lin)(sigmoid)(g)(res_lin)(r), inner sep=2.5pt] {};
\end{scope}

\end{tikzpicture}
}
\caption{\textbf{GatedLinear Architecture.} The premarket tensor is compressed via low-rank bilinear projection $\mathbf{Z} = \mathbf{V}^\top\mathbf{X}\mathbf{U}$ (only 2,428 parameters). A gating mechanism $\mathbf{g} \in [0,1]^{60}$ modulates the correction magnitude per time step. Despite 9$\times$ fewer parameters than AttnCorrect, this architecture achieves the best overall performance when combined with Random Forest.}
\label{fig:gated_arch}
\end{figure*}
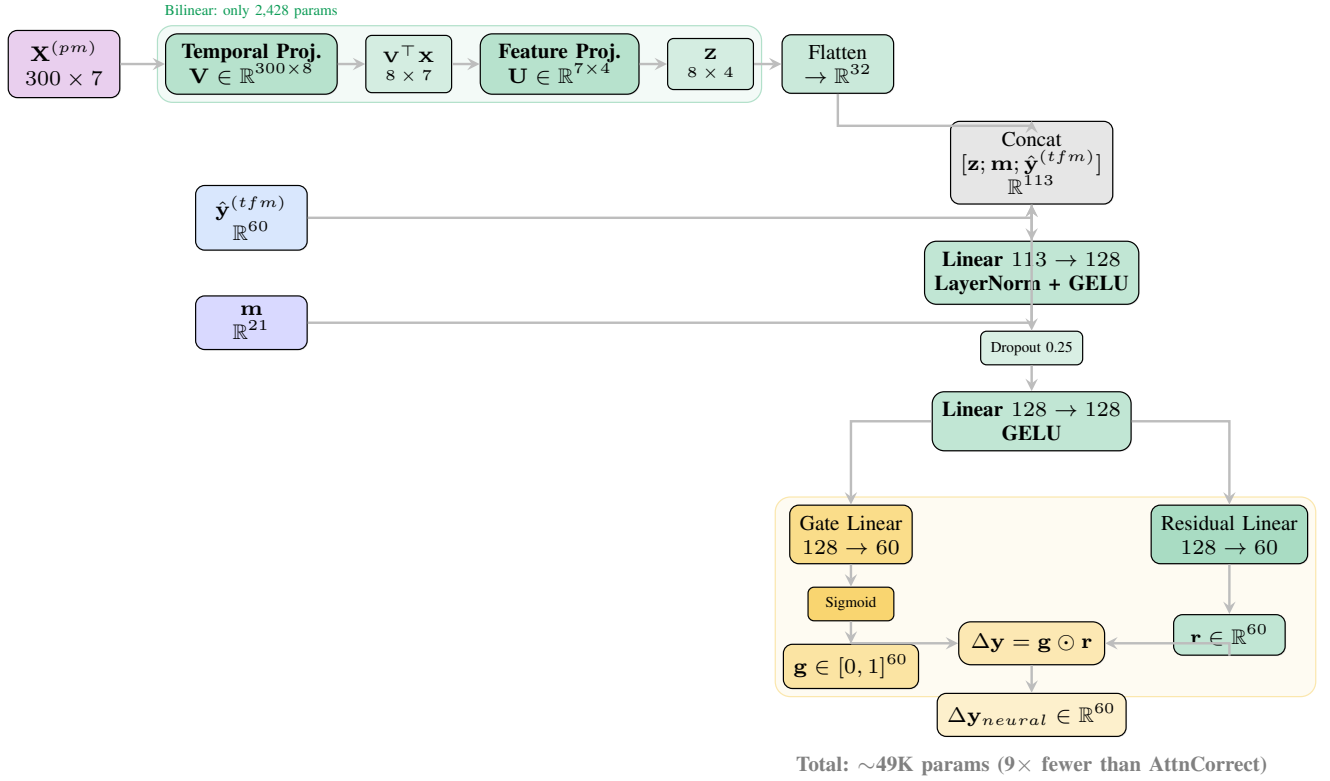

Figure \ref{fig:gated_arch} presents the GatedLinear architecture with only ${\sim}$49K trainable parameters---9$\times$ fewer than AttnCorrect.

\subsubsection{Low-Rank Bilinear Projection}

Instead of attention over the full sequence, we compress the premarket tensor through learned projection matrices:
\begin{equation}
    \mathbf{Z} = \mathbf{V}^\top \mathbf{X}^{(pm)} \mathbf{U} \in \mathbb{R}^{8 \times 4}
\end{equation}
where $\mathbf{V} \in \mathbb{R}^{300 \times 8}$ projects the temporal dimension and $\mathbf{U} \in \mathbb{R}^{7 \times 4}$ projects features. This bilinear form \cite{tenenbaum2000separating} captures interactions between temporal positions and feature channels with only $300 \times 8 + 7 \times 4 = 2,428$ parameters.

The temporal projection $\mathbf{V}$ is initialized with a slight linear trend to encourage learning of temporal patterns. The result $\mathbf{Z}$ is flattened to $\mathbf{z} \in \mathbb{R}^{32}$.

\subsubsection{Gated Correction}

Features are concatenated: $[\mathbf{z}; \mathbf{m}; \hat{\mathbf{y}}^{(tfm)}] \in \mathbb{R}^{113}$ and encoded through a two-layer MLP. The gating mechanism, inspired by LSTM \cite{hochreiter1997long} and GRU \cite{cho2014learning} gates, computes:
\begin{equation}
    \mathbf{g} = \sigma(\mathbf{W}_g \mathbf{h} + \mathbf{b}_g) \in [0,1]^{60}
\end{equation}
\begin{equation}
    \mathbf{r} = \mathbf{W}_r \mathbf{h} + \mathbf{b}_r \in \mathbb{R}^{60}
\end{equation}
\begin{equation}
    \Delta\mathbf{y}_{neural} = \mathbf{g} \odot \mathbf{r}
\end{equation}

The gate $\mathbf{g}$ controls correction magnitude per time step, allowing the model to selectively correct when confident.

\subsection{Training Details}

All experiments use a fixed random seed of 42 across PyTorch, NumPy, and Python's random module to ensure full reproducibility. Both models are trained with AdamW optimizer \cite{loshchilov2019decoupled}:
\begin{itemize}
    \item Learning rate: $5 \times 10^{-4}$ (AttnCorrect), $8 \times 10^{-4}$ (GatedLinear)
    \item Weight decay: $10^{-4}$
    \item Early stopping patience: 20 (AttnCorrect), 25 (GatedLinear)
    \item Gradient clipping: max norm 1.0
    \item Dropout: 0.3 (AttnCorrect), 0.25 (GatedLinear)
    \item Random seed: 42 (fixed for all random number generators)
\end{itemize}

\subsection{Ablation Study Design}

Table~\ref{tab:ablation_design} summarizes our comprehensive ablation with 12 model variants across 5 baselines, 2 full models, and 5 ablation variants.

\begin{table}[t]
\centering
\caption{Ablation Study Design: 12 Model Variants}
\label{tab:ablation_design}
\resizebox{\columnwidth}{!}{
\begin{tabular}{clll}
\toprule
\textbf{ID} & \textbf{Model} & \textbf{Ablated Component} & \textbf{Research Question} \\
\midrule
\multicolumn{4}{l}{\textit{Baselines (5 models)}} \\
01 & HistMean & -- & Naive baseline \\
02 & MLP & -- & Standard neural baseline \\
03 & LSTM & -- & Recurrent sequence modeling \\
04 & BiLSTM & -- & Bidirectional modeling \\
05 & TimesFM Base & -- & Frozen foundation model \\
\midrule
\multicolumn{4}{l}{\textit{Full Hybrid Models (2 models)}} \\
06 & AttnCorrect+RF & None (Full) & Attention + classical \\
07 & GatedLinear+RF & None (Full) & Bilinear + classical \\
\midrule
\multicolumn{4}{l}{\textit{Ablation Variants (5 models)}} \\
08 & AttnCorrect-NoRF & Random Forest & How much does RF add? \\
09 & GatedLinear-NoRF & Random Forest & How much does RF add? \\
10 & GatedLinear-NoGate & Gating mechanism & Is gating necessary? \\
11 & GatedLinear-NoBilinear & Bilinear projection & Is bilinear helpful? \\
12 & AttnCorrect-NoAttn & Self-attention & Is attention necessary? \\
\bottomrule
\end{tabular}
}
\end{table}

\section{Experimental Setup}

\subsection{Dataset}

We evaluate on 1-minute bar data for 10 major technology stocks representing diverse market capitalizations and volatility profiles. Table~\ref{tab:stock_info} provides dataset statistics. \rev{The timeframe spans December 2024 to January 2026.}

\begin{table}[t]
\centering
\caption{Dataset Statistics: 10 Technology Stocks}
\label{tab:stock_info}
\resizebox{\columnwidth}{!}{
\begin{tabular}{lcccc}
\toprule
\textbf{Stock} & \textbf{Train Days} & \textbf{Val Days} & \textbf{Test Days} & \textbf{Training Volatility} \\
\midrule
NVDA & 186 & 40 & 40 & 0.001627 \\
MSFT & 186 & 40 & 40 & 0.001000 \\
AAPL & 186 & 40 & 40 & 0.001098 \\
GOOG & 186 & 40 & 40 & 0.001050 \\
GOOGL & 186 & 40 & 40 & 0.001070 \\
AMZN & 186 & 40 & 40 & 0.001190 \\
META & 186 & 40 & 40 & 0.001298 \\
AVGO & 186 & 40 & 40 & 0.001764 \\
TSLA & 186 & 40 & 40 & 0.001988 \\
NFLX & 104 & 22 & 23 & 0.001150 \\
\midrule
\textbf{Total} & \multicolumn{4}{c}{\textbf{2,011,399 rows across 10 tickers}} \\
\bottomrule
\end{tabular}
}
\end{table}

The dataset spans approximately 266 trading days per stock (except NFLX with 149 days due to data availability). Stock-specific volatility varies significantly, from 0.001000 (MSFT, most stable) to 0.001988 (TSLA, most volatile).

\subsection{Data Leakage Prevention}

To ensure temporal validity and avoid look-ahead bias:

\textbf{1) Strict input/label time separation:} For each trading date, all model inputs use only pre-9:30 AM data (previous day's regular session + current day's premarket). Targets are computed exclusively from 9:30--10:30 AM.

\textbf{2) Chronological splits:} Data is split by trading day in strict chronological order---no shuffling. Test days occur strictly after all training and validation days.

\textbf{3) Training-only normalization:} Stock-specific volatility for feature normalization is computed using \textbf{training dates only}.

\textbf{4) Model fitting restricted to training set:} TimesFM remains frozen. Neural correctors and Random Forest are fitted only on training data; validation is used only for early stopping.

\subsection{Evaluation Metrics}

\rev{We report three complementary correlation metrics to provide a complete picture of predictive quality, alongside error metrics:}

\begin{itemize}
    \item \textbf{MAE (\%)}: Mean Absolute Error of return predictions
    \item \textbf{RMSE (\%)}: Root Mean Squared Error
    \item \rev{\textbf{Mean Per-Day Correlation ($\overline{\rho}_{day}$)}: For each test day $d$, we compute the Pearson correlation between the 60-bar predicted return vector $\hat{\mathbf{y}}_d$ and the 60-bar actual return vector $\mathbf{y}_d$, then average across all $D$ test days: $\overline{\rho}_{day} = \frac{1}{D}\sum_{d=1}^{D} \rho(\hat{\mathbf{y}}_d, \mathbf{y}_d)$. This measures \textit{within-day temporal alignment}---whether the model correctly predicts when returns are larger or smaller within each trading session.}
    \item \rev{\textbf{Cross-Day Correlation ($\rho_{cross}$)}: Pearson correlation between the cumulative predicted return per day and the cumulative actual return per day, computed across all $D$ test days. This measures whether the model correctly predicts which days have positive vs.\ negative net returns---i.e., \textit{directional forecasting across days}.}
    \item \rev{\textbf{Pooled Correlation ($\rho_{pool}$)}: Pearson correlation computed by concatenating all predictions and all actuals across all days and bars into single vectors.}
\end{itemize}

\section{Results}

\subsection{Aggregate Results}

Table~\ref{tab:main_results} presents aggregate results across all 10 stocks, sorted by RMSE. \rev{All three correlation metrics are reported.}

\begin{table}[t]
\centering
\caption{\rev{Aggregate Results Across 10 Stocks (Sorted by RMSE). Three correlation metrics: mean per-day ($\overline{\rho}_{day}$), cross-day ($\rho_{cross}$), and pooled ($\rho_{pool}$).}}
\label{tab:main_results}
\resizebox{\columnwidth}{!}{
\begin{tabular}{lccccc}
\toprule
\textbf{Model} & \textbf{MAE(\%)}$\downarrow$ & \textbf{RMSE(\%)}$\downarrow$ & \rev{$\overline{\rho}_{day}$}$\uparrow$ & \rev{$\rho_{cross}$}$\uparrow$ & \rev{$\rho_{pool}$}$\uparrow$ \\
\midrule
\rowcolor{positive!18}
\textbf{07\_GatedLinear+RF} & \textbf{\rev{0.1079}} & \textbf{\rev{0.1535}} & \textbf{\rev{0.3730}} & \rev{0.5631} & \textbf{\rev{0.5972}} \\
\rowcolor{positive!12}
06\_AttnCorrect+RF & \rev{0.1081} & \rev{0.1547} & \rev{0.3678} & \textbf{\rev{0.5819}} & \rev{0.5890} \\
\midrule
11\_GatedLinear-NoBilinear & \rev{0.1071} & \rev{0.1600} & \rev{0.3422} & \rev{0.5055} & \rev{0.5040} \\
10\_GatedLinear-NoGate & \rev{0.1084} & \rev{0.1666} & \rev{0.2864} & \rev{0.3620} & \rev{0.4317} \\
02\_MLP & \rev{0.1105} & \rev{0.1679} & \rev{0.2407} & \rev{0.4957} & \rev{0.4584} \\
09\_GatedLinear-NoRF & \rev{0.1087} & \rev{0.1710} & \rev{0.2147} & \rev{0.3058} & \rev{0.3219} \\
08\_AttnCorrect-NoRF & \rev{0.1091} & \rev{0.1740} & \rev{0.2335} & \rev{0.3829} & \rev{0.3698} \\
03\_LSTM & \rev{0.1082} & \rev{0.1744} & \rev{0.3519} & \rev{0.4628} & \rev{0.4943} \\
04\_BiLSTM & \rev{0.1090} & \rev{0.1779} & \rev{0.3420} & \rev{0.4653} & \rev{0.4737} \\
12\_AttnCorrect-NoAttn & \rev{0.1109} & \rev{0.1922} & \rev{0.1347} & \rev{0.4179} & \rev{0.1997} \\
05\_TimesFM Base & 0.1113 & 0.1946 & \rev{0.0586} & \rev{$-$0.0357} & 0.0614 \\
01\_HistMean & 0.1115 & 0.1946 & \rev{0.0442} & \rev{0.0000} & 0.0610 \\
\bottomrule
\end{tabular}
}
\end{table}

\textbf{Key Observations:}

\begin{enumerate}
    \item \textbf{Hybrid methods dramatically outperform all baselines}: \rev{Mean per-day correlation improves from 0.059 (TimesFM) to 0.373 (GatedLinear+RF)---a \textbf{6.4$\times$ improvement}. Pooled correlation reaches 0.597.}
    \item \textbf{GatedLinear+RF achieves best overall performance}: Best RMSE (\rev{0.1535}\%), best MAE (\rev{0.1079}\%), and \rev{best mean per-day (0.373) and pooled (0.597) correlation} with 9$\times$ fewer parameters
    \item \rev{\textbf{Cross-day vs.\ per-day correlations reveal different model strengths}: AttnCorrect+RF achieves the highest cross-day correlation (0.582), indicating slightly stronger daily directional forecasting, while GatedLinear+RF leads on per-day and pooled metrics.}
\end{enumerate}

\subsection{Per-Stock Performance Analysis}

Table~\ref{tab:per_stock} provides detailed per-stock results comparing the two full hybrid models against frozen TimesFM. \rev{We report RMSE, mean per-day correlation, and pooled correlation.}

\begin{table*}[t]
\centering
\caption{Per-Stock Performance: TimesFM vs.\ Hybrid Corrections. \rev{Mean per-day correlation ($\overline{\rho}_{day}$) and pooled correlation ($\rho_{pool}$) shown separately.}}
\label{tab:per_stock}
\resizebox{\textwidth}{!}{
\begin{tabular}{lccccccccc}
\toprule
& \multicolumn{3}{c}{\textbf{RMSE (\%)}} & \multicolumn{3}{c}{\rev{\textbf{Mean Per-Day Corr ($\overline{\rho}_{day}$)}}} & \multicolumn{3}{c}{\rev{\textbf{Pooled Corr ($\rho_{pool}$)}}} \\
\cmidrule(lr){2-4} \cmidrule(lr){5-7} \cmidrule(lr){8-10}
\textbf{Stock} & \textbf{TFM} & \textbf{Attn+RF} & \textbf{Gated+RF} & \textbf{TFM} & \textbf{Attn+RF} & \textbf{Gated+RF} & \rev{\textbf{TFM}} & \rev{\textbf{Attn+RF}} & \rev{\textbf{Gated+RF}} \\
\midrule
NVDA & 0.2466 & \rev{0.1689} & \rev{\textbf{0.1683}} & \rev{0.018} & \rev{\textbf{0.485}} & \rev{0.473} & \rev{0.165} & \rev{0.734} & \rev{\textbf{0.739}} \\
MSFT & 0.1228 & \rev{\textbf{0.0971}} & \rev{0.1010} & \rev{0.099} & \rev{\textbf{0.417}} & \rev{0.424} & \rev{0.075} & \rev{\textbf{0.633}} & \rev{0.624} \\
AAPL & 0.1095 & \rev{0.1073} & \rev{\textbf{0.1068}} & \rev{0.033} & \rev{0.135} & \rev{\textbf{0.171}} & \rev{0.040} & \rev{0.311} & \rev{\textbf{0.347}} \\
GOOG & 0.2140 & \rev{0.1659} & \rev{\textbf{0.1651}} & \rev{0.054} & \rev{\textbf{0.396}} & \rev{0.383} & \rev{$-$0.014} & \rev{0.644} & \rev{\textbf{0.646}} \\
GOOGL & 0.2151 & \rev{0.1690} & \rev{\textbf{0.1616}} & \rev{0.120} & \rev{0.376} & \rev{\textbf{0.408}} & \rev{0.070} & \rev{0.638} & \rev{\textbf{0.669}} \\
AMZN & 0.1463 & \rev{0.1272} & \rev{\textbf{0.1258}} & \rev{0.087} & \rev{0.344} & \rev{\textbf{0.351}} & \rev{0.085} & \rev{0.505} & \rev{\textbf{0.527}} \\
META & 0.1636 & \rev{0.1370} & \rev{\textbf{0.1363}} & \rev{0.083} & \rev{0.398} & \rev{\textbf{0.399}} & \rev{0.158} & \rev{\textbf{0.665}} & \rev{0.659} \\
AVGO & 0.3183 & \rev{0.2372} & \rev{\textbf{0.2320}} & \rev{0.010} & \rev{0.392} & \rev{\textbf{0.395}} & \rev{$-$0.075} & \rev{\textbf{0.688}} & \rev{0.685} \\
TSLA & 0.2541 & \rev{\textbf{0.1948}} & \rev{0.1954} & \rev{0.109} & \rev{\textbf{0.461}} & \rev{0.458} & \rev{0.136} & \rev{\textbf{0.647}} & \rev{0.644} \\
NFLX & 0.1558 & \rev{\textbf{0.1428}} & \rev{0.1431} & \rev{$-$0.027} & \rev{\textbf{0.274}} & \rev{0.268} & \rev{$-$0.025} & \rev{0.426} & \rev{\textbf{0.434}} \\
\midrule
\textbf{Average} & 0.1946 & \rev{0.1547} & \rev{\textbf{0.1535}} & \rev{0.059} & \rev{0.368} & \rev{\textbf{0.373}} & \rev{0.061} & \rev{0.589} & \rev{\textbf{0.597}} \\
\textbf{Best on} & 0/10 & \rev{3/10} & \rev{\textbf{7/10}} & 0/10 & \rev{4/10} & \rev{\textbf{6/10}} & 0/10 & \rev{4/10} & \rev{\textbf{6/10}} \\
\bottomrule
\end{tabular}
}
\end{table*}

\textbf{Per-Stock Insights:}

\begin{itemize}
    \item \textbf{GatedLinear+RF wins on \rev{7}/10 stocks for RMSE}, including the most volatile stocks (AVGO, TSLA)
    \item \rev{\textbf{Both methods transform near-zero per-day correlations into moderate-to-strong ones}}
    \item \textbf{Highest improvements on volatile stocks}: NVDA and AVGO show largest correlation gains
    \item \rev{\textbf{Per-day vs.\ pooled correlations differ}: For AAPL, per-day correlation is only 0.171 (GatedLinear+RF) while pooled is 0.347, suggesting the model captures cross-day variance structure better than within-day temporal patterns for less volatile stocks.}
\end{itemize}

\subsection{Ablation Analysis: Component Contributions}

\begin{table}[t]
\centering
\caption{Detailed Ablation Analysis: With vs.\ Without Each Component. \rev{We report mean per-day ($\overline{\rho}_{day}$) and cross-day ($\rho_{cross}$) correlation contributions.}}
\label{tab:ablation_detailed}
\resizebox{\columnwidth}{!}{
\begin{tabular}{lcccccc}
\toprule
\textbf{Component} & \textbf{RMSE$_{with}$} & \textbf{RMSE$_{w/o}$} & \rev{$\overline{\rho}_{day,with}$} & \rev{$\overline{\rho}_{day,w/o}$} & \rev{$\Delta\overline{\rho}_{day}$} & \rev{$\Delta\rho_{cross}$} \\
\midrule
\rowcolor{positive!20}
RF (GatedLinear) & \rev{0.1535} & \rev{0.1710} & \rev{0.3730} & \rev{0.2147} & \rev{\textcolor{positive}{\textbf{+0.1583}}} & \rev{\textcolor{positive}{\textbf{+0.2573}}} \\
\rowcolor{positive!20}
RF (AttnCorrect) & \rev{0.1547} & \rev{0.1740} & \rev{0.3678} & \rev{0.2335} & \rev{\textcolor{positive}{\textbf{+0.1343}}} & \rev{\textcolor{positive}{\textbf{+0.1990}}} \\
\rowcolor{positive!15}
Self-Attention & \rev{0.1740} & \rev{0.1922} & \rev{0.2335} & \rev{0.1347} & \rev{\textcolor{positive}{\textbf{+0.0988}}} & \rev{\textcolor{positive}{$-$0.0350}} \\
\midrule
\rowcolor{negative!10}
Gating Mechanism & \rev{0.1710} & \rev{0.1666} & \rev{0.2147} & \rev{0.2864} & \rev{\textcolor{negative}{$-$0.0717}} & \rev{\textcolor{negative}{$-$0.0562}} \\
\rowcolor{negative!15}
Bilinear Projection & \rev{0.1710} & \rev{0.1600} & \rev{0.2147} & \rev{0.3422} & \rev{\textcolor{negative}{\textbf{$-$0.1276}}} & \rev{\textcolor{negative}{$-$0.1997}} \\
\bottomrule
\end{tabular}
}
\vspace{0.3em}
\footnotesize{\rev{$\Delta\overline{\rho}_{day}$: positive means component improves mean per-day correlation; negative means removing component \textit{improves} performance. $\Delta\rho_{cross}$: same for cross-day correlation.}}
\end{table}

\subsubsection{Finding 1: Classical Residual Learning \rev{Provides the Largest Single-Component Contribution}}

\rev{\textbf{Random Forest provides the largest improvement of any individual component, whether neural or classical}:}

\begin{itemize}
    \item \textbf{GatedLinear}: RF adds \rev{+0.158 mean per-day correlation (0.215 $\to$ 0.373) and +0.257 cross-day correlation}, reduces RMSE by \rev{10.2}\%
    \item \textbf{AttnCorrect}: RF adds \rev{+0.134 mean per-day correlation (0.234 $\to$ 0.368) and +0.199 cross-day correlation}, reduces RMSE by \rev{11.1}\%
\end{itemize}

This finding demonstrates that classical machine learning methods remain highly valuable even when combined with neural approaches for foundation model adaptation. \rev{RF's contribution is particularly large for cross-day correlation, indicating it is effective at capturing features that predict daily directional returns.}

\subsubsection{Finding 2: Simpler Architectures Outperform Without Classical Components}

\textbf{The most surprising result}: When Random Forest is removed, simpler neural architectures outperform more complex ones.

Comparing GatedLinear-NoRF (\rev{0.215} \rev{per-day} corr) vs. GatedLinear-NoBilinear (\rev{0.342} \rev{per-day} corr):
\begin{itemize}
    \item Removing bilinear projection \textbf{improves} \rev{mean per-day} correlation by \rev{+0.128}
    \item Removing bilinear projection \textbf{improves} RMSE by \rev{6.4}\%
\end{itemize}

The bilinear projection compresses $300 \times 7 = 2,100$ dimensions to just 32---a 65$\times$ compression that loses fine-grained temporal information. The NoBilinear variant uses the last 60 premarket bars flattened directly (420 dimensions), preserving recent dynamics better.

\subsubsection{Finding 3: Self-Attention Provides \rev{the} Largest \rev{Positive} Neural Contribution}

Self-attention contributes \rev{+0.099 mean per-day} correlation---the largest \rev{positive} contribution from any \rev{neural-only} component:

\begin{itemize}
    \item \textbf{With attention}: RMSE=\rev{0.1740}\%, \rev{$\overline{\rho}_{day}$}=\rev{0.2335}
    \item \textbf{Without (mean pooling)}: RMSE=\rev{0.1922}\%, \rev{$\overline{\rho}_{day}$}=\rev{0.1347}
\end{itemize}

Self-attention enables flexible temporal pattern learning across the 300-bar premarket sequence, allowing the model to attend to relevant time periods dynamically. \rev{The cross-day correlation contribution is minimal ($-$0.035), confirming that self-attention's main value is in within-day temporal modeling rather than daily directional prediction.}

\subsubsection{Finding 4: Gating Mechanism Provides Marginal Benefit}

Removing the gating mechanism slightly \textbf{improves} performance:
\begin{itemize}
    \item \rev{$\Delta\overline{\rho}_{day}$} = \rev{$-$0.072} (removing gate improves by \rev{0.072})
    \item The learned gates may introduce unnecessary complexity without adding predictive value
\end{itemize}

\subsection{Parameter Efficiency Analysis}

Table~\ref{tab:params} details the parameter breakdown for both architectures.

\begin{table}[t]
\centering
\caption{Parameter Count Breakdown}
\label{tab:params}
\begin{tabular}{lrr}
\toprule
\textbf{Component} & \textbf{AttnCorrect} & \textbf{GatedLinear} \\
\midrule
Premarket processing & ${\sim}$66,000 & 2,428 \\
Attention/Bilinear & ${\sim}$200,000 & -- \\
Encoder/Fusion MLP & ${\sim}$190,000 & ${\sim}$31,000 \\
Output layers & ${\sim}$15,000 & ${\sim}$15,000 \\
\midrule
\textbf{Total trainable} & \textbf{${\sim}$471,000} & \textbf{${\sim}$49,000} \\
\textbf{Ratio} & 9.6$\times$ & 1$\times$ \\
\midrule
\textbf{Performance} & \rev{0.368/$\,$0.589} Corr & \textbf{\rev{0.373/$\,$0.597} Corr} \\
\bottomrule
\end{tabular}
\end{table}

GatedLinear achieves \textbf{better performance with 9.6$\times$ fewer parameters}, demonstrating that parameter efficiency does not require sacrificing accuracy in this domain.

\section{Discussion}

\rev{
\subsection{On Correlation Metrics for Multi-Step Forecasting}

A critical methodological point: for multi-step time series predictions (60 bars $\times$ $D$ days), how correlation is computed significantly affects reported results. Pooled correlation (flattening all predictions into one vector) can be inflated by cross-day variance structure---if the model merely captures that some days are more volatile than others, pooled correlation will be nonzero even without genuine within-day predictive ability. Mean per-day correlation is more conservative and more relevant for intraday trading, as it measures whether the model correctly predicts the \textit{temporal pattern} of returns within each session. Cross-day correlation captures a different skill: predicting which days will have positive vs.\ negative cumulative returns. We recommend reporting all three metrics in future work on multi-step financial forecasting.
}

\subsection{Why Does Classical Residual Learning \rev{Complement Neural Correction}?}

Several factors explain Random Forest's strong performance:

\begin{enumerate}
    \item \textbf{Complementary feature spaces}: RF operates on hand-crafted multiscale statistics (21 dimensions) capturing domain knowledge, while neural networks process raw sequences
    \item \textbf{Non-linear feature interactions}: Decision trees naturally capture complex interactions between features without explicit specification
    \item \textbf{Robustness to outliers}: Tree-based methods handle the heavy-tailed distributions common in financial data
    \item \textbf{Residual learning}: RF learns patterns that neural networks systematically miss, providing orthogonal improvements
\end{enumerate}

\subsection{Why Do Simpler Architectures Excel Without RF?}

The bilinear projection's 65$\times$ compression is too aggressive:

\begin{itemize}
    \item Loses fine-grained temporal dynamics important for prediction
    \item Forces the model to learn optimal temporal patterns that may not generalize
    \item Recent premarket information (last 60 bars) is more predictive than temporally-aggregated patterns
\end{itemize}

When RF is present, it compensates by accessing premarket summary statistics directly, explaining why the full GatedLinear+RF system achieves best performance despite the bilinear bottleneck.

\subsection{Practical Recommendations}

Based on our comprehensive ablation:

\begin{enumerate}
    \item \textbf{Always include classical residual learning}: RF provides \rev{contributions nearly matching} at minimal computational cost
    \item \textbf{Start with simpler neural architectures}: Complex neural components may not provide benefits proportional to their parameter cost
    \item \textbf{Self-attention is valuable for sequence modeling}: Provides largest \rev{positive} neural contribution (+\rev{0.099 per-day} correlation)
    \item \textbf{Avoid aggressive dimensionality reduction}: Preserve recent temporal information unless classical components can compensate
    \item \textbf{Prioritize parameter efficiency}: GatedLinear+RF achieves best results with 9$\times$ fewer parameters
    \item \rev{\textbf{Report multiple correlation metrics}: }
\end{enumerate}

\subsection{Limitations and Future Work}

\textbf{Limitations:}
\begin{itemize}
    \item Evaluation limited to 10 large-cap technology stocks
    \item Test period spans 40 days; longer evaluation needed
    \item Transaction costs and market impact not modeled
    \item Only TimesFM tested; other foundation models may behave differently
    \item \rev{The limited baseline correlation of frozen TimesFM means that even modest absolute improvements yield large relative gains}
    \item \rev{Premarket signals may be inherently more predictive of early-session returns than of later trading hours, so the reported correlations may not generalize to full-day forecasting}
    \item \rev{This work evaluates the hybrid correction methodology rather than claiming that TimesFM itself is suited for financial prediction tasks}
\end{itemize}

\textbf{Future directions:}
\begin{itemize}
    \item Extend to other asset classes and market conditions
    \item Investigate adaptive weighting of neural vs. classical components
    \item Apply to other foundation models (Chronos, Lag-Llama)
    \item Develop theoretically-grounded guidelines for neural-classical integration
\end{itemize}

\section{Conclusion}

We presented a comprehensive study of hybrid neural-classical correction for adapting frozen time series foundation models to high-frequency stock prediction. Through systematic ablation across 10 technology stocks and 12 model variants, we reveal that:

\begin{enumerate}
    \item \textbf{Hybrid approaches achieve dramatic improvements}: \rev{6.4$\times$ mean per-day correlation improvement over frozen TimesFM (0.059 $\to$ 0.373), with pooled correlation reaching 0.597}
    \item \textbf{Classical and neural components contribute nearly equally}
    \item \textbf{Simpler neural architectures outperform complex ones} when classical components are removed
    \item \textbf{Parameter efficiency is achievable}: GatedLinear+RF achieves best performance with 9$\times$ fewer neural parameters
    
\end{enumerate}

Our key message: \textbf{effective foundation model adaptation requires thoughtful integration of neural and classical methods}. Classical machine learning remains highly valuable even in the era of foundation models, providing complementary capabilities that neural networks alone cannot match.

\section*{Acknowledgment}

This paper was prepared with the assistance of AI tools. \rev{Specifically, AI tools were used for text editing, citation formatting and optimization assistance, and GitHub Copilot and other tools were used for coding assistance.} The authors take full responsibility for the content and have verified all AI-assisted contributions.

\rev{The authors thank the anonymous reviewers for their valuable feedback.}

\bibliographystyle{IEEEtran}

\end{document}